%% file: ms.tex
\documentclass[10pt,twocolumn,letterpaper]{article}

\usepackage{iccv}
\usepackage{times}
\usepackage{epsfig}
\usepackage{graphicx}
\usepackage{appendix}
\usepackage{amsmath,amssymb}
\usepackage{subfigure}

\usepackage{xcolor}
\usepackage{multirow}
\usepackage{caption}
\usepackage[breaklinks=true,bookmarks=false]{hyperref}

 \iccvfinalcopy 

\def\iccvPaperID{936} 

\ificcvfinal\fi
\begin{document}

\title{Learning Efficient Convolutional Networks through Network Slimming}

\author{Zhuang Liu$^1$\thanks{This work was done when Zhuang Liu and Zhiqiang Shen were interns at Intel Labs China. Jianguo Li is the corresponding author.}
\quad
Jianguo Li$^2$
\quad
Zhiqiang Shen$^3$
\quad
Gao Huang$^4$
\quad
Shoumeng Yan$^2$
\quad
Changshui Zhang$^1$\\
$^1$Tsinghua University\quad
$^2$Intel Labs China\quad
$^3$Fudan University\quad
$^4$Cornell University\\
{\tt\small \{liuzhuangthu, zhiqiangshen0214\}@gmail.com, }{\tt\small \{jianguo.li, shoumeng.yan\}@intel.com, }\\{\tt\small gh349@cornell.edu, zcs@mail.tsinghua.edu.cn}}

\maketitle
\input{macro}

\thispagestyle{empty}
\begin{abstract}
\input{abstract}
\end{abstract}

\input{intro}
\input{related.tex}

\input{method}

\input{experiment}

\input{analysis}

\section{Conclusion}
We proposed the network slimming technique to learn more compact CNNs. It directly imposes sparsity-induced regularization on the scaling factors in batch normalization layers, and unimportant channels can thus be automatically identified during training and then pruned. On multiple datasets, we have shown that the proposed method is able to significantly decrease the computational cost (up to 20$\times$) of state-of-the-art networks, with no accuracy loss. More importantly, the proposed method simultaneously reduces the model size, run-time memory, computing operations while introducing minimum overhead to the training process, and the resulting models require no special libraries/hardware for efficient inference.

\vspace{10pt}
\noindent\textbf{Acknowledgements}. Gao Huang is supported by the International Postdoctoral Exchange Fellowship Program of China Postdoctoral Council (No.20150015).
Changshui Zhang is supported by NSFC and DFG joint project NSFC 61621136008/DFG TRR-169.

{\small
\bibliographystyle{ieee}
\bibliography{example_paper}
}

\newpage
\input{sup}

\end{document}

%% file: macro.tex
\newcommand{\scf}{scaling factor }
\newcommand{\scfs}{scaling factors }
\newcommand{\suf}{-SC}

\newcommand{\zl}[1]{\textcolor{blue}{[ZL: #1]}}
\newcommand{\jl}[1]{\textcolor{red}{[JL: #1]}}
\newcommand{\hl}[1]{\textcolor{blue}{#1}}

\newcommand\blfootnote[1]{%
  \begingroup
  \renewcommand\thefootnote{}\footnote{#1}%
  \addtocounter{footnote}{-1}%
  \endgroup
}

%% file: abstract.tex

The deployment of deep convolutional neural networks (CNNs) in many real world applications is largely hindered by their high computational cost. In this paper, we propose a novel learning scheme for CNNs to simultaneously 1) reduce the model size; 2) decrease the run-time memory footprint; and 3) lower the number of computing operations, without compromising accuracy. This is achieved by enforcing channel-level sparsity in the network in a simple but effective way. Different from many existing approaches, the proposed method directly applies to modern CNN architectures, introduces minimum overhead to the training process, and requires no special software/hardware accelerators for the resulting models. We call our approach \emph{network slimming}, which takes wide and large networks as input models, but during training insignificant channels are automatically identified and pruned afterwards, yielding thin and compact models with comparable accuracy. We empirically demonstrate the effectiveness of our approach with several state-of-the-art CNN models, including VGGNet, ResNet and DenseNet, on various image classification datasets. For VGGNet, a multi-pass version of network slimming gives a 20$\times$ reduction in model size and a 5$\times$ reduction in computing operations.


%% file: intro.tex
\section{Introduction}

\begin{figure*}[]
\centering
\includegraphics[width=0.95\linewidth]{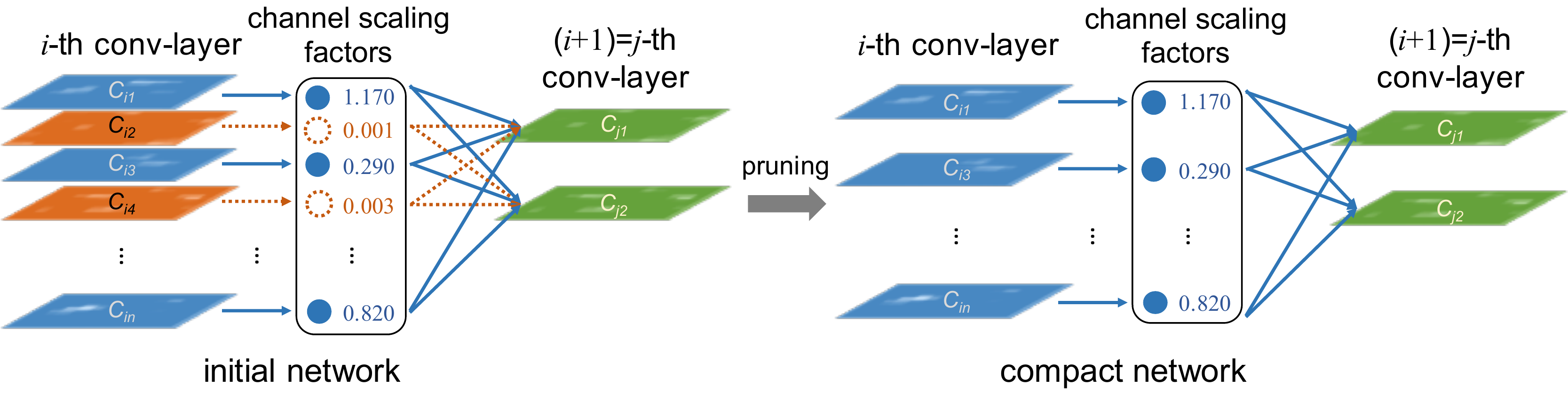}
\vskip -0.1in
\caption{
We associate a scaling factor (reused from a batch normalization layer) with each channel in convolutional layers.
Sparsity regularization is imposed on these scaling factors during training to automatically identify unimportant channels. The channels with small scaling factor values (in orange color) will be pruned (left side). After pruning, we obtain compact models (right side), which are then fine-tuned to achieve comparable (or even higher) accuracy as normally trained full network.}
\label{scaling}
\vskip -0.15in
\end{figure*}

In recent years, convolutional neural networks (CNNs) have become the dominant approach for a variety of computer vision tasks, e.g., image classification \cite{alexnet}, object detection \cite{rcnn}, semantic segmentation \cite{fcn}.
Large-scale datasets, high-end modern GPUs and new network architectures allow the development of unprecedented large CNN models.
For instance, from AlexNet \cite{alexnet}, VGGNet \cite{vgg} and GoogleNet \cite{googlenet} to ResNets \cite{resnet}, the ImageNet Classification Challenge winner models have evolved from 8 layers to more than 100 layers.

However, larger CNNs, although with stronger representation power, are more resource-hungry. For instance, a 152-layer ResNet \cite{resnet} has more than 60 million parameters and requires more than 20 Giga float-point-operations (FLOPs) when inferencing an image with resolution 224$\times$ 224. This is unlikely to be affordable on resource constrained platforms such as mobile devices, wearables or Internet of Things (IoT) devices.

The deployment of CNNs in real world applications are mostly constrained by
\textit{1) Model size}: CNNs' strong representation power comes from their millions of trainable parameters. Those parameters, along with network structure information, need to be stored on disk and loaded into memory during inference time. As an example, storing a typical CNN trained on ImageNet consumes more than 300MB space, which is a big resource burden to embedded devices.
\textit{2) Run-time memory}: During inference time, the intermediate activations/responses of CNNs could even take more memory space than storing the model parameters, even with batch size 1. This is not a problem for high-end GPUs, but unaffordable for many applications with low computational power.
\textit{3) Number of computing operations:} The convolution operations are computationally intensive on high resolution images. A large CNN may take several minutes to process one single image on a mobile device, making it unrealistic to be adopted for real applications.



Many works have been proposed to compress large CNNs or directly learn more efficient CNN models for fast inference. These include low-rank approximation \cite{svd}, network quantization \cite{hashnet, han-learning} and binarization \cite{xnor-net, binarynet}, weight pruning \cite{han-learning}, dynamic inference \cite{msdnet}, etc.
However, most of these methods can only address one or two challenges mentioned above.
Moreover, some of the techniques require specially designed software/hardware accelerators for execution speedup \cite{xnor-net, binarynet,han-learning}.

Another direction to reduce the resource consumption of large CNNs is to sparsify the network.
Sparsity can be imposed on different level of structures \cite{ggsparse,lessismore,ssl,group-sparse,scnn},
which yields considerable model-size compression and inference speedup.
However, these approaches generally require special software/hardware accelerators to harvest the gain in memory or time savings, though it is easier than non-structured sparse weight matrix as in \cite{han-learning}.

In this paper, we propose \emph{network slimming}, a simple yet effective network training scheme, which addresses all the aforementioned challenges when deploying large CNNs under limited resources. Our approach imposes L1 regularization on the scaling factors in batch normalization (BN) layers, thus it is easy to implement without introducing any change to existing CNN architectures. Pushing the values of BN scaling factors towards zero with L1 regularization enables us to identify insignificant channels (or neurons), as each  scaling factor corresponds to a specific convolutional channel (or a neuron in a fully-connected layer). This facilitates the channel-level pruning at the followed step. The additional regularization term rarely hurt the performance. In fact, in some cases it leads to higher generalization accuracy. Pruning unimportant channels may sometimes temporarily degrade the performance, but this effect can be compensated by the followed fine-tuning of the pruned network.
After pruning, the resulting narrower network is much more compact in terms of model size, run-time memory, and computing operations compared to the initial wide network. The above process can be repeated for several times, yielding a multi-pass network slimming scheme which leads to even more compact network.


Experiments on several benchmark datasets and different network architectures show that we can obtain CNN models with up to 20x mode-size compression and 5x reduction in computing operations of the original ones, while achieving the same or even higher accuracy.
Moreover, our method achieves model compression and inference speedup with conventional hardware and deep learning software packages,
since the resulting narrower model is free of any  sparse storing format or computing operations.

%% file: related.tex
\section{Related Work}
In this section, we discuss related work from five aspects.

\vspace{5pt}
\noindent\textbf{Low-rank Decomposition} approximates weight matrix in neural networks with low-rank matrix using techniques like Singular Value Decomposition (SVD) \cite{svd}.
This method works especially well on fully-connected layers, yielding $\sim$3x model-size compression however without notable speed acceleration, since computing operations in CNN mainly come from convolutional layers.

\vspace{5pt}
\noindent\textbf{Weight Quantization.} HashNet \cite{hashnet} proposes to quantize the network weights. Before training, network weights are hashed to different groups and within each group weight the value is shared. In this way only the shared weights and hash indices need to be stored, thus a large amount of storage space could be saved. \cite{han-learning} uses a improved quantization technique in a deep compression pipeline and achieves 35x to 49x compression rates on AlexNet and VGGNet. However, these techniques can neither save run-time memory nor inference time, since during inference shared weights need to be restored to their original positions.

\cite{xnor-net, binarynet} quantize real-valued weights into binary/ternary weights (weight values restricted to $\{-1, 1\}$ or $\{-1, 0, 1\}$).
This yields a large amount of model-size saving, and significant speedup could also be obtained given bitwise operation libraries. However, this aggressive low-bit approximation method usually comes with a moderate accuracy loss.

\vspace{5pt}
\noindent\textbf{Weight Pruning / Sparsifying.}
\cite{han-learning} proposes to prune the unimportant connections with small weights in  trained neural networks. The resulting network's weights are mostly zeros thus the storage space can be reduced by storing the model in a sparse format. However, these methods can only achieve speedup with dedicated sparse matrix operation libraries and/or hardware. The run-time memory saving is also very limited since most memory space is consumed by the activation maps (still dense) instead of the weights.

In \cite{han-learning}, there is no guidance for sparsity during training.
\cite{training-sparse} overcomes this limitation by explicitly imposing sparse constraint over each weight with additional gate variables,
 and achieve high compression rates by pruning connections with zero gate values.
This method achieves better compression rate than \cite{han-learning}, but suffers from the same drawback.

\vspace{5pt}
\noindent\textbf{Structured Pruning / Sparsifying.}
Recently, \cite{pruning} proposes to prune channels with small incoming weights in trained CNNs,
and then fine-tune the network to regain accuracy.
\cite{ggsparse} introduces sparsity by random deactivating input-output channel-wise connections in convolutional layers before training,
which also yields smaller networks with moderate accuracy loss.
Compared with these works, we explicitly impose channel-wise sparsity in the optimization objective during training,
leading to smoother channel pruning process and little accuracy loss.

\cite{lessismore} imposes neuron-level sparsity during training thus some neurons could be pruned to obtain compact networks. \cite{ssl} proposes a Structured Sparsity Learning (SSL) method to sparsify different level of structures (e.g. filters, channels or layers) in CNNs. Both methods utilize group sparsity regualarization during training to obtain structured sparsity. Instead of resorting to group sparsity on convolutional weights, our approach imposes simple L1 sparsity on channel-wise scaling factors, thus the optimization objective is much simpler.

Since these methods prune or sparsify part of the network structures (e.g., neurons, channels) instead of individual weights, they usually require less specialized libraries (e.g. for sparse computing operation) to achieve inference speedup and run-time memory saving. Our network slimming also falls into this category, with absolutely no special libraries needed to obtain the benefits.

\vspace{5pt}
\noindent\textbf{Neural Architecture Learning.} While state-of-the-art CNNs are typically designed by experts \cite{alexnet, vgg, resnet}, there are also some explorations on automatically learning network architectures.
\cite{subsupnn} introduces sub-modular/super-modular optimization for network architecture search with a given resource budget.
Some recent works \cite{rl1,rl2} propose to learn neural architecture automatically with reinforcement learning.
The searching space of these methods are extremely large, thus one needs to train hundreds of models to distinguish good from bad ones. Network slimming can also be treated as an approach for architecture learning, despite the choices are  limited to the width of each layer. However, in contrast to the aforementioned methods, network slimming learns network architecture through only a single training process, which is in line with our goal of efficiency.

%% file: method.tex
\section{Network slimming}
We aim to provide a simple scheme to achieve channel-level sparsity in deep CNNs. In this section, we first discuss the advantages and challenges of channel-level sparsity, and introduce how we leverage the scaling layers in batch normalization to effectively identify and prune unimportant channels in the network.

\vspace{5pt}
\noindent\textbf{Advantages of Channel-level Sparsity.} As discussed in prior works \cite{ssl,pruning,deep-compression},  sparsity can be realized at different levels, e.g., weight-level, kernel-level, channel-level or layer-level. Fine-grained level (e.g., weight-level) sparsity gives the highest flexibility and generality leads to higher compression rate, but it usually requires special software or hardware accelerators to do fast inference on the sparsified model \cite{deep-compression}. On the contrary, the coarsest layer-level sparsity does not require special packages to harvest the inference speedup, while it is less flexible as some whole layers need to be pruned. In fact, removing layers is only effective when the depth is sufficiently large, e.g., more than 50 layers \cite{ssl,stochastic}. In comparison, channel-level sparsity provides a nice tradeoff between flexibility and ease of implementation. It can be applied to any typical CNNs or fully-connected networks (treat each neuron as a channel), and the resulting network is essentially a ``thinned'' version of the unpruned network, which can be efficiently inferenced on conventional CNN platforms.

\vspace{5pt}
\noindent\textbf{Challenges.} Achieving channel-level sparsity requires pruning all the incoming and outgoing connections associated with a channel. This renders the method of directly pruning weights on a pre-trained model ineffective, as it is unlikely that all the weights at the input or output end of a channel happen to have near zero values. As reported in \cite{pruning}, pruning channels on pre-trained ResNets can only lead to a reduction of $\sim$10\% in the number of parameters without suffering from accuracy loss. \cite{ssl} addresses this problem by enforcing sparsity regularization into the training objective. Specifically, they adopt \emph{group LASSO} to push all the filter weights corresponds to the same channel towards zero simultaneously during training. However, this approach requires computing the gradients of the additional regularization term with respect to all the filter weights, which is nontrivial. We introduce a simple idea to address the above challenges, and the details are presented below.

\vspace{5pt}
\noindent\textbf{Scaling Factors and Sparsity-induced Penalty.}
Our idea is introducing a scaling factor $\gamma$ for each channel, which is multiplied to the output of that channel. Then we jointly train the network weights and these scaling factors, with sparsity regularization imposed on the latter. Finally we prune those channels with small factors, and fine-tune the pruned network. Specifically, the training objective of our approach is given by
\begin{equation}\label{lambda}
		L = \sum_{(x, y)}{l(f(x, W), y)} + \lambda \sum_{\gamma\in {\Gamma}} {g(\gamma)}
\end{equation}
where $(x,y)$ denote the train input and target, $W$ denotes the trainable weights, the first sum-term corresponds to the normal training loss of a CNN, $g(\cdot)$ is a sparsity-induced penalty on the scaling factors, and $\lambda$ balances the two terms. In our experiment, we choose $g(s) = |s|$, which is known as L1-norm and widely used to achieve sparsity. Subgradient descent is adopted as the optimization method for the non-smooth L1 penalty term. An alternative option is to replace the L1 penalty with the smooth-L1 penalty \cite{smoothL1} to avoid using sub-gradient at non-smooth point.

As pruning a channel essentially corresponds to removing all the incoming and outgoing connections of that channel, we can directly obtain a narrow network (see \autoref{scaling}) without resorting to any special sparse computation packages. The scaling factors act as the agents for channel selection. As they are jointly optimized with the network weights, the network can automatically identity insignificant channels, which can be safely removed without greatly affecting the generalization performance.


\begin{figure}[]
\centering
\centerline{\includegraphics[width=1.0\linewidth]{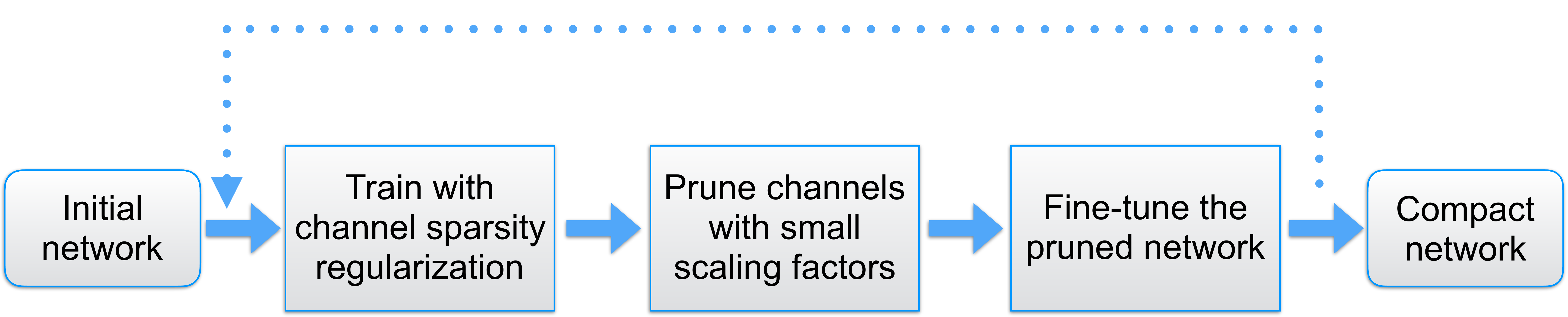}}
\caption{Flow-chart of network slimming procedure. The dotted-line is for the multi-pass/iterative scheme.}
\label{iterative}
\vskip -0.15in
\end{figure}

\vspace{5pt}
\noindent\textbf{Leveraging the Scaling Factors in BN Layers.}
Batch normalization \cite{bn} has been adopted by most modern CNNs as a standard approach to achieve fast convergence and better generalization performance. The way BN normalizes the activations motivates us to design a simple and efficient method to incorporates the channel-wise scaling factors.
Particularly, BN layer normalizes the internal activations using mini-batch statistics.
Let $z_{in}$ and $z_{out}$ be the input and output of a BN layer, $\mathcal{B}$ denotes the current mini-batch, BN layer performs the following transformation:
\begin{gather}
	\hat{z} = \frac{z_{in} - \mu_\mathcal{B}}{\sqrt{\sigma_\mathcal{B}^2 + \epsilon}}; \ \
	z_{out} = \gamma \hat{z} + \beta
\end{gather}
where $\mu_\mathcal{B}$ and $\sigma_\mathcal{B}$ are the mean and standard deviation values of input activations over $\mathcal{B}$,
$\gamma$ and $\beta$ are trainable affine transformation parameters (scale and shift) which provides the possibility of linearly transforming normalized activations back to any scales.

It is common practice to insert a BN layer after a convolutional layer, with channel-wise scaling/shifting parameters.
Therefore, we can directly leverage the $\gamma$ parameters in BN layers as the scaling factors we need for network slimming. It has the great advantage of introducing no overhead to the network.
In fact, this is perhaps also the most effective way we can learn meaningful scaling factors for channel pruning.
\textit{1)}, if we add scaling layers to a CNN without BN layer, the value of the scaling factors are not meaningful for evaluating the importance of a channel, because both convolution layers and scaling layers are linear transformations. One can obtain the same results by decreasing the scaling factor values while amplifying the weights in the convolution layers.
\textit{2)}, if we insert a scaling layer before a BN layer, the scaling effect of the scaling layer will be completely canceled by the normalization process in BN.
\textit{3)}, if we insert scaling layer after BN layer, there are two consecutive scaling factors for each channel.

\vspace{5pt}
\noindent\textbf{Channel Pruning and Fine-tuning.}
After training under channel-level sparsity-induced regularization, we obtain a model in which many scaling factors are near zero (see \autoref{scaling}).
Then we can prune channels with near-zero scaling factors, by removing all their incoming and outgoing connections and corresponding weights.
We prune channels with a global threshold across all layers, which is defined as a certain percentile of all the scaling factor values.
For instance, we prune 70\% channels with lower scaling factors by choosing the percentile threshold as 70\%.
By doing so, we obtain a more compact network with less parameters and run-time memory, as well as less computing operations.

Pruning may temporarily lead to some accuracy loss, when the pruning ratio is high.
But this can be largely compensated by the followed fine-tuning process on the pruned network.
In our experiments, the fine-tuned narrow network can even achieve higher accuracy than the original unpruned network in many cases.

\vspace{5pt}
\noindent\textbf{Multi-pass Scheme.}
We can also extend the proposed method from single-pass learning scheme (training with sparsity regularization, pruning, and fine-tuning) to a multi-pass scheme. Specifically, a network slimming procedure results in a narrow network, on which we could again apply the whole training procedure to learn an even more compact model.
This is illustrated by the dotted-line in \autoref{iterative}. Experimental results show that this multi-pass scheme can lead to even better results in terms of compression rate.

\vspace{5pt}
\noindent\textbf{Handling Cross Layer Connections and Pre-activation Structure.}
The network slimming process introduced above can be directly applied to most plain CNN architectures such as AlexNet \cite{alexnet} and VGGNet \cite{vgg}. While some adaptations are required when it is applied to modern networks with \emph{cross layer connections} and the \emph{pre-activation} design such as ResNet \cite{identity} and DenseNet \cite{densenet}. For these networks, the output of a layer may be treated as the input of multiple subsequent layers, in which a BN layer is placed before the convolutional layer. In this case, the sparsity is achieved at the incoming end of a layer, i.e., the layer selectively uses a subset of channels it received. To harvest the parameter and computation savings at test time, we need to place a \emph{channel selection} layer
to mask out insignificant channels we have identified.

%% file: experiment.tex
\input{table}

\section{Experiments}

We empirically demonstrate the effectiveness of
network slimming on several benchmark datasets. We implement our method based on the publicly available Torch \cite{torch} implementation for ResNets by \cite{fb}. The code is available at \url{https://github.com/liuzhuang13/slimming}.

\subsection{Datasets}
\vspace{2pt}
\noindent\textbf{CIFAR.}
      The two CIFAR datasets \cite{cifar} consist of natural images with resolution 32$\times$32. CIFAR-10 is drawn from 10 and CIFAR-100 from 100 classes. The train and test sets contain 50,000 and 10,000 images respectively. On CIFAR-10, a validation set of 5,000 images is split from the training set for the search of $\lambda$ (in \autoref{lambda}) on each model. We report the final test errors after training or fine-tuning on all training images. A standard data augmentation scheme (shifting/mirroring) \cite{resnet, stochastic, netinnet} is adopted. The input data is normalized using channel means and standard deviations. We also compare our method with \cite{pruning} on CIFAR datasets.

\vspace{5pt}
\noindent\textbf{SVHN.}
The Street View House Number (SVHN) dataset \cite{svhn} consists of 32x32 colored digit images. Following common practice \cite{maxout, stochastic, netinnet} we use all the 604,388 training images, from which we split a validation set of 6,000 images for model selection during training. The test set contains 26,032 images. During training, we select the model with the lowest validation error as the model to be pruned (or the baseline model). We also report the test errors of the models with lowest validation errors during fine-tuning.

\vspace{5pt}
\noindent\textbf{ImageNet.}
The ImageNet dataset contains 1.2 million training images and 50,000 validation images of 1000 classes. We adopt the data augmentation scheme as in \cite{fb}.
We report the single-center-crop validation error of the final model.

\vspace{5pt}
\noindent\textbf{MNIST.}
MNIST is a handwritten digit dataset containing 60,000 training images and 10,000 test images. To test the effectiveness of our method on a fully-connected network (treating each neuron as a channel with 1$\times$1 spatial size), we compare our method with \cite{ssl} on this dataset.

\subsection{Network Models}
On CIFAR and SVHN dataset, we evaluate our method on three popular network architectures: VGGNet\cite{vgg}, ResNet \cite{resnet} and DenseNet \cite{densenet}. The VGGNet is originally designed for ImageNet classification. For our experiment a variation of the original VGGNet for CIFAR dataset is taken from \cite{vgggithub}. For ResNet, a 164-layer pre-activation  ResNet with bottleneck structure (ResNet-164) \cite{identity} is used. For DenseNet, we use a 40-layer DenseNet with growth rate 12 (DenseNet-40).

On ImageNet dataset, we adopt the 11-layer (8-conv + 3 FC) ``VGG-A'' network \cite{vgg} model with batch normalization from \cite{alexnet-github}. We remove the dropout layers since we use relatively heavy data augmentation. To prune the neurons in fully-connected layers, we treat them as convolutional channels with 1$\times$1 spatial size. 

On MNIST dataset, we evaluate our method on the same 3-layer fully-connected network as in \cite{ssl}.

\subsection{Training, Pruning and Fine-tuning}
\noindent\textbf{Normal Training.} We train all the networks normally from scratch as baselines. All the networks are trained using SGD. On CIFAR and SVHN datasets we train using mini-batch size 64 for 160 and 20 epochs, respectively. The initial learning rate is set to 0.1, and is divided by 10 at 50\% and 75\% of the total number of training epochs. On ImageNet and MNIST datasets, we train our models for 60 and 30 epochs respectively, with a batch size of 256, and an initial learning rate of 0.1 which is divided by 10 after 1/3 and 2/3 fraction of training epochs.  We use a weight decay of $10^{-4}$ and a Nesterov momentum \cite{nesterov} of 0.9 without dampening. The weight initialization introduced by \cite{init} is adopted. Our optimization settings closely follow the original implementation at \cite{fb}. In all our experiments, we initialize all channel scaling factors to be 0.5, since this gives higher accuracy for the baseline models compared with default setting (all initialized to be 1) from \cite{fb}.

\vspace{5pt}
\noindent\textbf{Training with Sparsity.} For CIFAR and SVHN datasets, when training with channel sparse regularization, the hyper-parameteer $\lambda$, which controls the tradeoff between empirical loss and sparsity, is determined by a grid search over 10$^{-3}$, 10$^{-4}$, 10$^{-5}$ on CIFAR-10 validation set. For VGGNet we choose $\lambda$=10$^{-4}$ and for ResNet and DenseNet $\lambda$=10$^{-5}$. For VGG-A on ImageNet, we set $\lambda$=10$^{-5}$. All other settings are kept the same as in normal training.

\vspace{5pt}
\noindent\textbf{Pruning.} When we prune the channels of models trained with sparsity, a pruning threshold on the \scfs needs to be determined. Unlike in \cite{pruning} where different layers are pruned by different ratios, we use a global pruning threshold for simplicity. The pruning threshold is determined by a percentile among all \scfs, e.g., 40\% or 60\% channels are pruned. The pruning process is implemented by building a new narrower model and copying the corresponding weights from the model trained with sparsity.

\vspace{5pt}
\noindent\textbf{Fine-tuning.} After the pruning we obtain a narrower and more compact model, which is then fine-tuned. On CIFAR, SVHN and MNIST datasets, the fine-tuning uses the same optimization setting as in training. For ImageNet dataset, due to time constraint, we fine-tune the pruned VGG-A with a learning rate of 10$^{-3}$ for only 5 epochs.
\subsection{Results}
\begin{figure}[]
\centering
\centerline{\includegraphics[width=0.9\columnwidth]{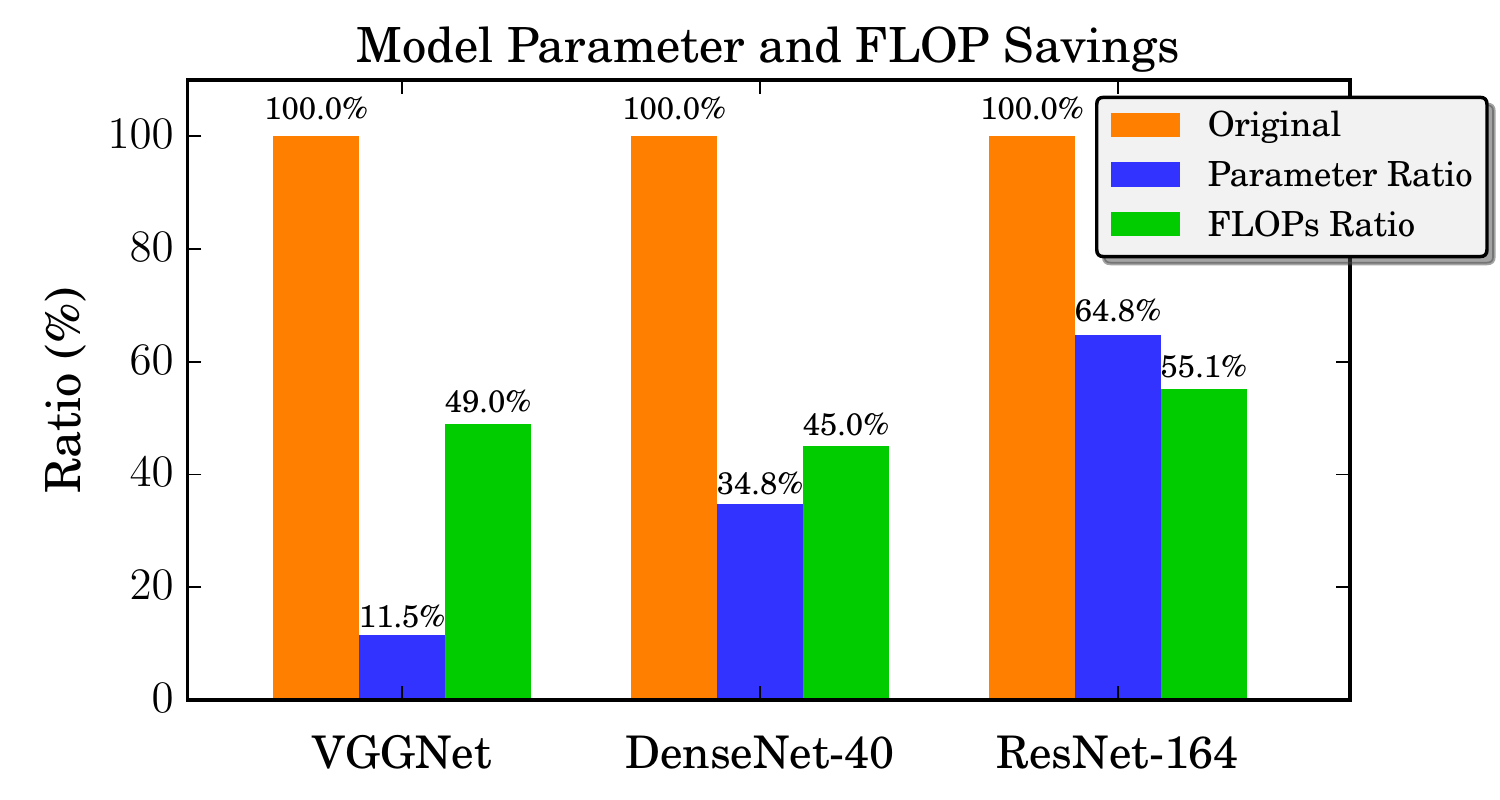}}
\vskip -0.05in
\caption{Comparison of pruned models with \textbf{lower} test errors on CIFAR-10 than the original models. The blue and green bars are parameter and FLOP ratios between pruned and original models.}
\label{bar}
\vskip -0.15in
\end{figure}

\noindent\textbf{CIFAR and SVHN}
The results on CIFAR and SVHN
are shown in \autoref{results}. We mark all lowest test errors of a model in \textbf{boldface}.

\vspace{5pt}
\noindent\textbf{Parameter and FLOP reductions.}
The purpose of network slimming is to reduce the amount of computing resources needed. The last row of each model has $\ge$ 60\% channels pruned while still maintaining similar accuracy to the baseline. The parameter saving can be up to 10$\times$. The FLOP reductions are typically around $50\%$. To highlight network slimming's efficiency, we plot the resource savings in \autoref{bar}. It can be observed that VGGNet has a large amount of redundant parameters that can be pruned. On ResNet-164 the parameter and FLOP savings are relatively insignificant, we conjecture this is due to its ``bottleneck'' structure has already  functioned as selecting channels.  Also, on CIFAR-100 the reduction rate is typically slightly lower than CIFAR-10 and SVHN, which is possibly due to the fact that CIFAR-100 contains more classes.

\vspace{5pt}
\noindent\textbf{Regularization Effect.}
From \autoref{results}, we can observe that, on ResNet and DenseNet, typically when $40\%$ channels are pruned, the fine-tuned network can achieve a lower test error than the original models. For example, DenseNet-40 with 40\% channels pruned achieve a test error of 5.19\% on CIFAR-10, which is almost 1\% lower than the original model. We hypothesize this is due to the regularization effect of L1 sparsity on channels, which naturally provides feature selection in intermediate layers of a network. We will analyze this effect in the next section.

\begin{table}[]
\small
\centering
\setlength{\tabcolsep}{1em}
\begin{tabular}{c|c|c}
\hline
VGG-A         & Baseline           & 50\% Pruned        \\ \hline
Params           & 132.9M              & 23.2M              \\
Params Pruned           & -                  & 82.5\%             \\ \hline
FLOPs            & 4.57$\times$10$^{10}$ & 3.18$\times$10$^{10}$ \\
FLOPs Pruned           & -                  & 30.4\%             \\ \hline
Validation Error (\%) & 36.69             & 36.66             \\ \hline
\end{tabular}
\caption{Results on ImageNet.}
\label{imagenet}
\end{table}

\vspace{6pt}
\noindent\textbf{ImageNet.}
The results for ImageNet dataset are summarized in \autoref{imagenet}.
When 50\% channels are pruned, the parameter saving is more than 5$\times$, while the FLOP saving is only 30.4\%. This is due to the fact that only 378 (out of 2752) channels from all the computation-intensive convolutional layers are pruned, while 5094 neurons (out of 8192) from the parameter-intensive fully-connected layers are pruned. It is worth noting that our method can achieve the savings with no accuracy loss on the 1000-class ImageNet dataset, where other methods for efficient CNNs \cite{ggsparse, pruning, ssl, xnor-net} mostly report accuracy loss.

\vspace{6pt}
\noindent\textbf{MNIST.}
On MNIST dataset, we compare our method with the Structured Sparsity Learning (SSL) method \cite{ssl} in \autoref{mnist}. Despite our method is mainly designed to prune channels in convolutional layers, it also works well in pruning neurons in fully-connected layers. In this experiment, we observe that pruning with a global threshold sometimes completely removes a layer, thus we prune 80\% of the neurons in each of the two intermediate layers. Our method slightly outperforms \cite{ssl}, in that a slightly lower test error is achieved while pruning more parameters.

\vspace{6pt}
We provide some additional experimental results in the supplementary materials, including 
(1) detailed structure of a compact VGGNet on CIFAR-10;
(2) wall-clock time and run-time memory savings in practice.
(3) comparison with a previous channel pruning method \cite{pruning};

\begin{table}[]
\centering
\small
\setlength{\tabcolsep}{0.3em}
\begin{tabular}{l|c|c|c}
\hline
Model        & Test Error (\%)    & Params Pruned  & \#Neurons        \\ \hline
Baseline       & 1.43   & -   & 784-500-300-10                \\ \hline
Pruned \cite{ssl} & 1.53  & 83.5\% & 434-174-78-10             \\ \hline
Pruned (ours)  & 1.49  & 84.4\%     & 784-100-60-10         \\ \hline
\end{tabular}
\caption{Results on MNIST.}
\label{mnist}
\end{table}

\input{iterative_table}
\input{save_table}
\begin{figure*}[]
\centering
\centerline{\includegraphics[width=1.90\columnwidth]{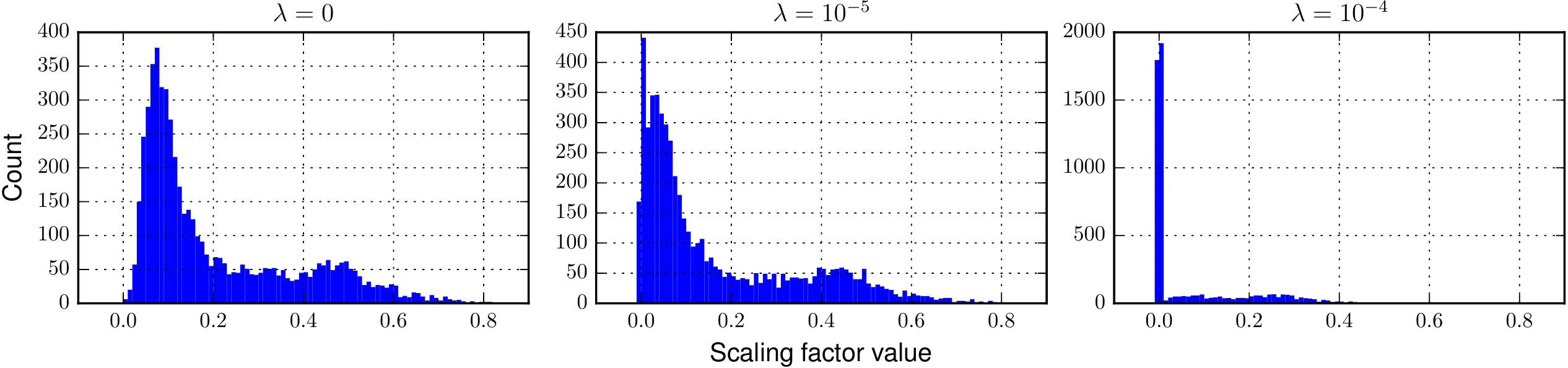}}
\vspace{-8pt}
\caption{Distributions of scaling factors in a trained VGGNet under various degree of sparsity regularization (controlled by the parameter $\lambda$). With the increase of $\lambda$, scaling factors become sparser.}
\label{dist}
\vskip -0.2in
\end{figure*}

\subsection{Results for Multi-pass Scheme}
We employ the multi-pass scheme on CIFAR datasets using VGGNet. Since there are no skip-connections, pruning away a whole layer will completely destroy the models. Thus, besides setting the percentile threshold as 50\%, we also put a constraint that at each layer, at most 50\% of channels can be pruned.

The test errors of models in each iteration are shown in \autoref{iterative_results}. As the pruning process goes, we obtain more and more compact models. On CIFAR-10, the trained model achieves the lowest test error in iteration 5. This model achieves 20$\times$ parameter reduction and 5$\times$ FLOP reduction, while still achieving \textbf{lower} test error. 
On CIFAR-100, after iteration 3, the test error begins to increase. This is possibly due to that it contains more classes than CIFAR-10, so pruning channels too agressively will inevitably hurt the performance. However, we can still prune near 90\% parameters and near 70\% FLOPs without notable accuracy loss.

%% file: table.tex
\setlength{\tabcolsep}{0.75em}
\begin{table*}[]
\small
  \centering
  \vspace{-10pt}
  \subtable[Test Errors on CIFAR-10]{
\begin{tabular}{l|ccccc}
\hline
\multicolumn{1}{c|}{Model}       & Test error (\%)    & Parameters & Pruned  & FLOPs              & Pruned  \\ \hline
VGGNet (Baseline)                     & 6.34          & 20.04M     & -      & 7.97$\times$10$^8$ & -      \\
VGGNet (70\% Pruned)                  & \textbf{6.20} & 2.30M      & 88.5\% & 3.91$\times$10$^8$ & 51.0\% \\ \hline
DenseNet-40 (Baseline)             & 6.11          & 1.02M      & -      & 5.33$\times$10$^8$ & -      \\
DenseNet-40 (40\% Pruned)          & \textbf{5.19} & 0.66M      & 35.7\% & 3.81$\times$10$^8$ & 28.4\% \\
DenseNet-40 (70\% Pruned)    & 5.65          & 0.35M      & 65.2\% & 2.40$\times$10$^8$ & 55.0\% \\ \hline
ResNet-164 (Baseline)              & 5.42          & 1.70M      & -      & 4.99$\times$10$^8$ & -      \\
ResNet-164 (40\% Pruned) & \textbf{5.08} & 1.44M      & 14.9\% & 3.81$\times$10$^8$ & 23.7\% \\
ResNet-164 (60\% Pruned)           & 5.27          & 1.10M      & 35.2\% & 2.75$\times$10$^8$ & 44.9\% \\ \hline
\end{tabular}
\label{70}
  }
  \vspace{-1ex}
  \subtable[Test Errors on CIFAR-100]{
\begin{tabular}{l|ccccc}
\hline
\multicolumn{1}{c|}{Model}        & Test error (\%)    & Parameters & Pruned  & FLOPs              & Pruned  \\ \hline
VGGNet (Baseline)                      & 26.74          & 20.08M     & -      & 7.97$\times$10$^8$ & -      \\
VGGNet (50\% Pruned)                   & \textbf{26.52}          & 5.00M      & 75.1\% & 5.01$\times$10$^8$ & 37.1\% \\ \hline
DenseNet-40 (Baseline)              & 25.36          & 1.06M      & -      & 5.33$\times$10$^8$ & -      \\
DenseNet-40 (40\% Pruned) & \textbf{25.28} & 0.66M      & 37.5\% & 3.71$\times$10$^8$ & 30.3\% \\
DenseNet-40 (60\% Pruned)           & 25.72          & 0.46M      & 54.6\% & 2.81$\times$10$^8$ & 47.1\% \\ \hline
ResNet-164 (Baseline)               & 23.37          & 1.73M      & -      & 5.00$\times$10$^8$ & -      \\
ResNet-164 (40\% Pruned)         & \textbf{22.87}          & 1.46M      & 15.5\% & 3.33$\times$10$^8$ & 33.3\% \\
ResNet-164 (60\% Pruned)                   & 23.91          & 1.21M      & 29.7\% & 2.47$\times$10$^8$ & 50.6\% \\ \hline
\end{tabular}
  }
  \vspace{-1ex}
    \subtable[Test Errors on SVHN]{
\begin{tabular}{l|ccccc}
\hline
\multicolumn{1}{c|}{Model}       & Test Error (\%)    & Parameters & Pruned  & FLOPs              & Pruned  \\ \hline
VGGNet (Baseline)                     & 2.17          & 20.04M     & -      & 7.97$\times$10$^8$ & -      \\
VGGNet (60\% Pruned)                  & \textbf{2.06} & 3.04M      & 84.8\% & 3.98$\times$10$^8$ & 50.1\% \\ \hline
DenseNet-40 (Baseline)             & 1.89          & 1.02M      & -      & 5.33$\times$10$^8$ & -      \\
DenseNet-40 (40\% Pruned)          & \textbf{1.79} & 0.65M      & 36.3\% & 3.69$\times$10$^8$ & 30.8\% \\
DenseNet-40 (60\% Pruned)    & 1.81          & 0.44M      & 56.6\% & 2.67$\times$10$^8$ & 49.8\% \\ \hline
ResNet-164 (Baseline)              & \textbf{1.78}          & 1.70M      & -      & 4.99$\times$10$^8$ & -      \\
ResNet-164 (40\% Pruned) & 1.85 & 1.46M      & 14.5\% & 3.44$\times$10$^8$ & 31.1\% \\
ResNet-164 (60\% Pruned)           & 1.81          & 1.12M      & 34.3\% & 2.25$\times$10$^8$ & 54.9\% \\ \hline
\end{tabular}
  }
\vskip -0.08in
 \caption{Results on CIFAR and SVHN datasets. ``Baseline'' denotes normal training without sparsity regularization. In column-1, ``60\% pruned'' denotes the fine-tuned model with 60\% channels pruned from the model trained with sparsity, etc. The pruned ratio of parameters and FLOPs are also shown in column-4\&6. Pruning a moderate amount (40\%) of channels can mostly lower the test errors. The accuracy could typically be maintained with $\ge$ 60\% channels pruned.}
 \label{results}
\vskip -0.15in
\end{table*}

%% file: iterative_table.tex
\setlength{\tabcolsep}{0.2em}
\begin{table}[]
\small
  \centering
  \subtable[Multi-pass Scheme on CIFAR-10]{
  \def\arraystretch{1}%
\begin{tabular}{l|cc|c|cc}
\hline
\multicolumn{1}{c|}{Iter} & Trained & Fine-tuned   & Params Pruned & FLOPs Pruned \\ \hline

1           & 6.38     & 6.51   & 66.7\%  & 38.6\% \\ \hline
2           & 6.23     & 6.11   & 84.7\%  & 52.7\% \\ \hline
3           & 5.87     & 6.10   &  91.4\%  & 63.1\% \\ \hline
4           & 6.19     & 6.59   &  95.6\%  & 77.2\% \\ \hline
5           & 5.96     & 7.73   &  98.3\%  & 88.7\% \\ \hline
6           & 7.79     & 9.70   &  99.4\%  & 95.7\% \\ \hline
\end{tabular}
  }

  \subtable[Multi-pass Scheme on CIFAR-100]{
\begin{tabular}{l|cc|c|cc}
\hline
\multicolumn{1}{c|}{Iter} & Trained & Fine-tuned   & Params Pruned &FLOPs Pruned\\ \hline

1           & 27.72     & 26.52   & 59.1\%  & 30.9\% \\ \hline
2           & 26.03     & 26.52   & 79.2\%  & 46.1\% \\ \hline
3           & 26.49     & 29.08   & 89.8\%  & 67.3\% \\ \hline
4           & 28.17     & 30.59   &  95.3\%  & 83.0\% \\ \hline
5           & 30.04     & 36.35   &  98.3\%  & 93.5\% \\ \hline
6           & 35.91     & 46.73   &  99.4\%  & 97.7\% \\ \hline
\end{tabular}
  }
 \vskip -0.07in  
 \caption{Results for multi-pass scheme on CIFAR-10 and CIFAR-100 datasets, using VGGNet. The baseline model has test errors of 6.34\% and 26.74\%. ``Trained'' and ``Fine-tuned'' columns denote the test errors (\%) of the model trained with sparsity, and the fine-tuned model after channel pruning, respectively. The parameter and FLOP pruned ratios correspond to the fine-tuned model in that row and the trained model in the next row.}
 \label{iterative_results}
 \vskip -0.15in
\end{table}

%% file: save_table.tex

%% file: analysis.tex
\begin{figure}[]
\vskip 0.0in
\centering
\hskip -0.15in
\centerline{\includegraphics[width=0.85\columnwidth]{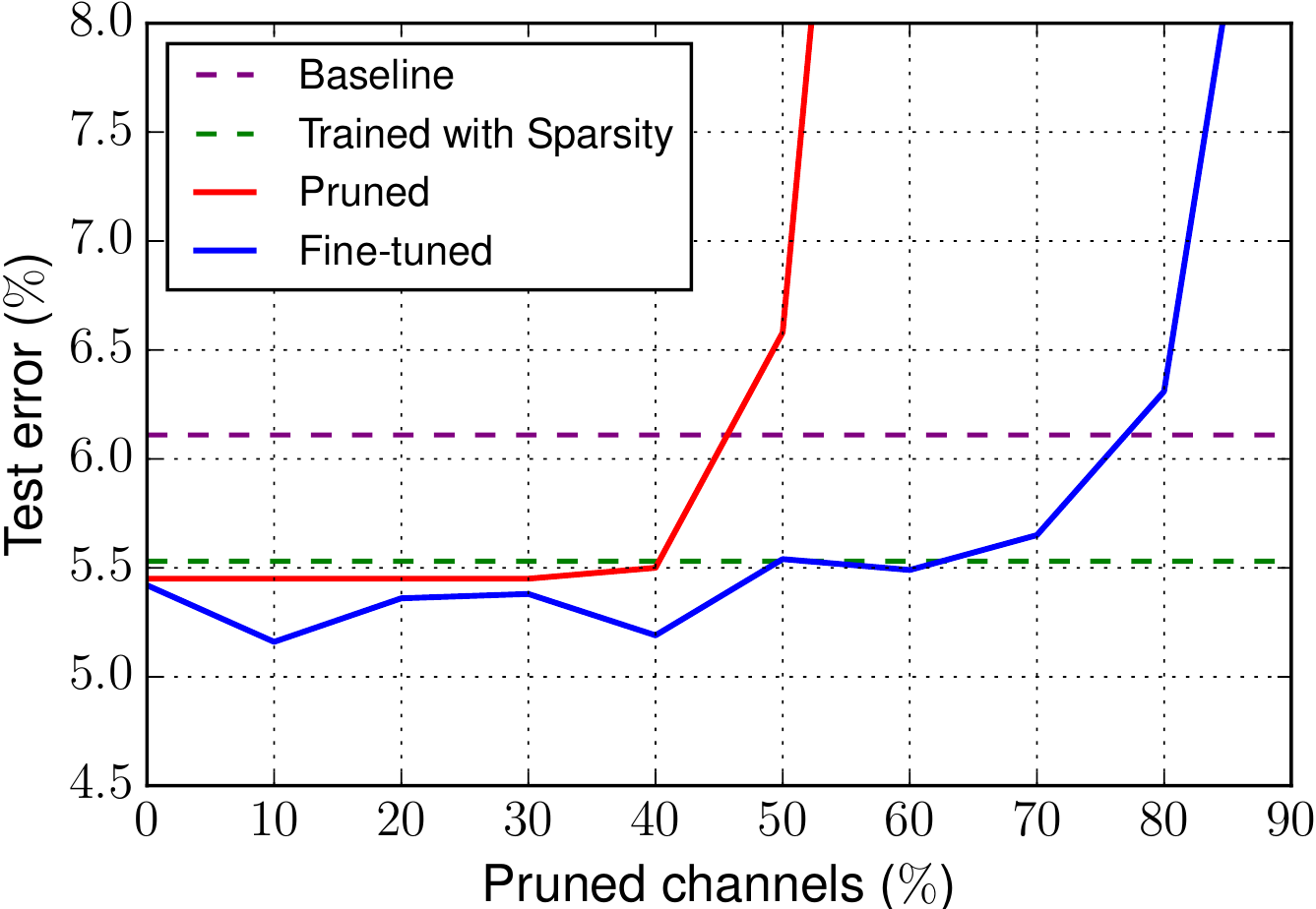}}
\caption{The effect of pruning varying percentages of channels, from DenseNet-40 trained on CIFAR-10 with $\lambda$=10$^{-5}$.}
\label{h}
\end{figure}

\section{Analysis}
There are two crucial hyper-parameters in network slimming, the pruned percentage $t$ and the coefficient of the sparsity regularization term $\lambda$ (see \autoref{lambda}). In this section, we analyze their effects in more detail.

\vspace{5pt}
\noindent\textbf{Effect of Pruned Percentage.}
Once we obtain a model trained with sparsity regularization, we need to decide what percentage of channels to prune from the model. If we prune too few channels, the resource saving can be very limited. However, it could be destructive to the model if we prune too many channels, and it may not be possible to recover the accuracy by fine-tuning. We train a DenseNet-40 model with $\lambda$=10$^{-5}$ on CIFAR-10 to show the effect of pruning a varying percentage of channels. The results are summarized in \autoref{h}.

From \autoref{h}, it can be concluded that the classification performance of the pruned or fine-tuned models degrade only when the pruning ratio surpasses a threshold. The fine-tuning process can typically compensate the possible accuracy loss caused by pruning. Only when the threshold goes beyond 80\%, the test error of fine-tuned model falls behind the baseline model. Notably, when trained with sparsity, even without fine-tuning, the model performs better than the original model. This is possibly due the the regularization effect of L1 sparsity on channel scaling factors.

\vspace{5pt}
\noindent\textbf{Channel Sparsity Regularization.}
The purpose of the L1 sparsity term is to force many of the scaling factors to be near zero. The parameter $\lambda$ in \autoref{lambda} controls its significance compared with the normal training loss. In \autoref{dist} we plot the distributions of scaling factors in the whole network with different $\lambda$ values. For this experiment we use a VGGNet trained on CIFAR-10 dataset.

\begin{figure}[]
\vspace{2pt}
\centering
\hskip -0.15in
\centerline{\includegraphics[width=0.9\columnwidth]{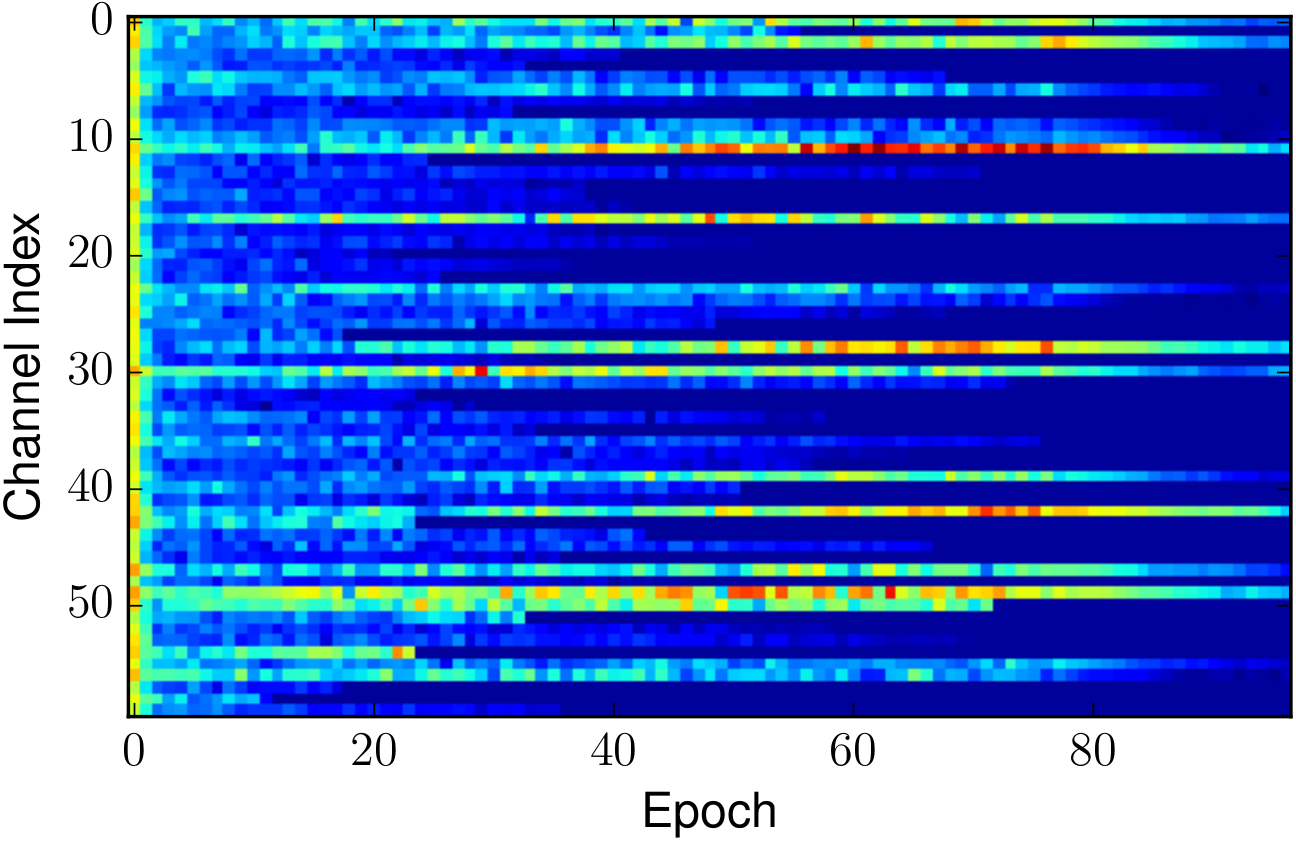}}
\caption{Visulization of channel scaling factors' change in scale along the training process, taken from the 11th conv-layer in VGGNet trained on CIFAR-10. Brighter color corresponds to larger value. The bright lines indicate the ``selected'' channels, the dark lines indicate channels that can be pruned.}
\label{heatmap}
\end{figure}

It can be observed that with the increase of $\lambda$, the scaling factors are more and more concentrated near zero. When $\lambda$=0, i.e., there's no sparsity regularization, the distribution is relatively flat.  When $\lambda$=10$^{-4}$, almost all scaling factors fall into a small region near zero. This process can be seen as a feature selection happening in intermediate layers of deep networks, where only channels with non-negligible scaling factors are chosen. We further visualize this process by a heatmap. \autoref{heatmap} shows the magnitude of scaling factors from one layer in VGGNet,  along the training process. Each channel starts with equal weights; as the training progresses, some channels' scaling factors become larger (brighter) while others become smaller (darker).

%% file: sup.tex
{\noindent \Large \bf Supplementary Materials}
\begin{appendices}	

\section{Detailed Structure of a Compact Network}
We show a detailed structure of a compact VGGNet on CIFAR-10 dataset in \autoref{save_table}. The compact model used is from the multi-pass scheme experiment (``Iter 5 Trained'' from \autoref{iterative_results} \textcolor{black}{(a)}). We observe that deeper layers tend to have more channels pruned. 

\setlength{\tabcolsep}{0.5em}
\begin{table}[h]
\small
\centering
  \def\arraystretch{1.1}%
\begin{tabular}{c|ccc|c}
\hline
Layer & Width & Width* & Pruned & P/F Pruned   \\ \hline1 & 64    & 22     & 65.6\% & 34.4\%        \\ 
2 & 64    & 62     & 3.1\% & 66.7\%        \\ 
3 & 128    & 83     & 35.2\% & 37.2\%        \\ 
4 & 128    & 119     & 7.0\% & 39.7\%        \\ 
5 & 256    & 193     & 24.6\% & 29.9\%        \\ 
6 & 256    & 168     & 34.4\% & 50.5\%        \\ 
7 & 256    & 85     & 66.8\% & 78.2\%        \\ 
8 & 256    & 40     & 84.4\% & 94.8\%        \\ 
9 & 512    & 32     & 93.8\% & 99.0\%        \\ 
10 & 512    & 32     & 93.8\% & 99.6\%        \\ 
11 & 512    & 32     & 93.8\% & 99.6\%        \\ 
12 & 512    & 32     & 93.8\% & 99.6\%        \\ 
13 & 512    & 32     & 93.8\% & 99.6\%        \\ 
14 & 512    & 32     & 93.8\% & 99.6\%        \\ 
15 & 512    & 32     & 93.8\% & 99.6\%        \\ 
16 & 512    & 38     & 92.6\% & 99.6\%        \\ 
 \hline Total & 5504  & 1034   & 81.2\% & 95.6\%/77.2\% \\\hline
\end{tabular}
\caption{Detailed structure of a compact VGGNet. ``Width'' and ``Width*''
denote each layer's number of channels in the original VGGNet (test error 6.34\%) and a compact VGGNet (test error 5.96\%) respectively. ``P/F Pruned'' denotes the parameter/FLOP pruned ratio at each layer.}
\label{save_table}
\end{table}

\section{Wall-clock Time and Run-time Memory Savings}
We test the wall-clock speed and memory footprint of a ``70\% pruned'' VGGNet (\autoref{results} \textcolor{black}{(a)}) on CIFAR-10 during inference time. The experiment is conducted using Torch \cite{torch} on a NVIDIA GeForce 1080 GPU with batch size 64. The result is shown in \autoref{wall}.

\setlength{\tabcolsep}{0.5em}
\begin{table}[]
\small
\centering
\label{my-label}
  \def\arraystretch{1.1}%
\begin{tabular}{l|c|c|c}
\hline
        VGGNet             & Time/Iter & Memory & Test Error (\%) \\ \hline 
Baseline    & 0.009s         & 697MB           &  6.34    \\ \hline
70\% Pruned & 0.005s         & 499MB           & 6.20     \\ \hline
\end{tabular}
\caption{Wall-clock time and run-time memory savings of a compact VGGNet.}
\label{wall}
\end{table}
The wall-clock time saving of this model roughly matches the FLOP saving shown in \autoref{results} \textcolor{black}{(a)}, despite the memory saving is not as significant. This is due to the fact that deeper layers, which have smaller activation maps and occupy less memory, tend to have more channels pruned, as shown by \autoref{save_table}. Note that all savings require no special libraries/hardware.

\section{Comparison with \cite{pruning}}
On CIFAR-10 and CIFAR-100 datasets, we compare our method with a previous channel pruning technique \cite{pruning}. Unlike network slimming which prunes channels with a global pruning threshold, \cite{pruning} prunes different pre-defined portion of channels at different layers. To make a comparison, we adopt the pruning criterion introduced in \cite{pruning} and closely follow the per-layer pruning strategy of \cite{pruning} on VGGNet \cite{vgggithub}. The result is shown in \autoref{compare}. Compared with \cite{pruning}, network slimming yields significantly lower test error with a similar compression rate.

\setlength{\tabcolsep}{0.5em}
\begin{table}[h]
\small
  \centering
  \subtable[CIFAR-10]{
  \def\arraystretch{1.1}%
\begin{tabular}{l|c|c}
\hline
\multicolumn{1}{c|} {Model} & Test Error (\%)   & Params Pruned \\ \hline
Baseline         & 6.34    & - \\ \hline
Pruned (\cite{pruning})        & 6.88    & 88.5\% \\ \hline
Pruned (ours)    & 6.20  & 88.5\% \\ \hline
\end{tabular}
  }
  
  \subtable[CIFAR-100]{
  \def\arraystretch{1.1}%
\begin{tabular}{l|c|c}
\hline
\multicolumn{1}{c|}{Model} & Test Error (\%)   & Params Pruned \\ \hline
Baseline          & 26.74    & - \\ \hline
Pruned (\cite{pruning})         & 28.36    & 76.0\% \\ \hline
Pruned (ours)        & 26.52  & 75.1\% \\ \hline
\end{tabular}
  }

 \vskip -0.07in  
 \caption{Comparison between our method and \cite{pruning}.}
 \label{compare}
 \vskip -0.15in
\end{table}

\end{appendices}